\documentclass{article}
\usepackage{arxiv}
\usepackage[utf8]{inputenc} % allow utf-8 input
\usepackage[T1]{fontenc}    % use 8-bit T1 fonts
\usepackage{hyperref}       % hyperlinks
\usepackage{url}            % simple URL typesetting
\usepackage{booktabs}       % professional-quality tables
\usepackage{amsfonts}       % blackboard math symbols
\usepackage{nicefrac}       % compact symbols for 1/2, etc.
\usepackage{microtype}      % microtypography
\usepackage{lipsum}         % Can be removed after putting your text content
\usepackage{graphicx}
\usepackage{natbib}
\usepackage{doi}
\usepackage{amsmath}
\usepackage{multirow} 

\title{DepFlow: Disentangled Speech Generation to Mitigate Semantic Bias in Depression Detection}

\newif\ifuniqueAffiliation
\uniqueAffiliationtrue

\ifuniqueAffiliation % Standard variant of author block
\author{
	Yuxin Li\textsuperscript{1*}, 
	Xiangyu Zhang\textsuperscript{2*},
	Yifei Li\textsuperscript{1}, 
	Zhiwei Guo\textsuperscript{1},
	Haoyang Zhang\textsuperscript{3},
	Eng Siong Chng\textsuperscript{1},
	Cuntai Guan\textsuperscript{1,4}\\
	\\
	\textsuperscript{1}College of Computing and Data Science, Nanyang Technological University, Singapore\\
	\textsuperscript{2}School of Electrical Engineering and Telecommunications, UNSW, Sydney, Australia\\
	\textsuperscript{3}School of Software and Microelectronics, Peking University, Beijing, China\\
	\textsuperscript{4}Center for AI in Medicine (C-AIM), Lee Kong Chian School of Medicine, NTU, Singapore\\
	\small{\textsuperscript{*}Equal Contribution}\\
	\small{\texttt{\{yuxin.li, YIFEI024, ZHIWEI004, ASESChng, ctguan\}@ntu.edu.sg}}\\
	\small{\texttt{xiangyu.zhang2@unsw.edu.au; zhang.haoyoung@stu.pku.edu.cn}}
}
\else
\usepackage{authblk}

\author[1]{Yuxin Li\textsuperscript{*}}
\author[2]{Xiangyu Zhang\textsuperscript{*}}
\author[1]{Yifei Li}
\author[1]{Zhiwei Guo}
\author[3]{Haoyang Zhang}
\author[1]{Eng Siong Chng}
\author[1,4]{Cuntai Guan}
\affil[1]{College of Computing and Data Science, Nanyang Technological University, Singapore}
\affil[2]{School of Electrical Engineering and Telecommunications, UNSW, Sydney, Australia}
\affil[3]{School of Software and Microelectronics, Peking University, Beijing, China}
\affil[4]{Center for AI in Medicine (C-AIM), Lee Kong Chian School of Medicine, NTU, Singapore}
\affil[ ]{\textsuperscript{*}Equal Contribution}
\fi

\renewcommand{\shorttitle}{DepFlow: Disentangled Speech Generation to Mitigate Semantic Bias in Depression Detection}

\hypersetup{
	pdftitle={DepFlow: Disentangled Speech Generation to Mitigate Semantic Bias in Depression Detection},
	pdfsubject={Speech Processing, Depression Detection, Deep Learning},
	pdfauthor={Yuxin Li, Xiangyu Zhang, Yifei Li, Zhiwei Guo, Haoyang Zhang, Eng Siong Chng, Cuntai Guan},
	pdfkeywords={Depression Detection, Speech Generation, Semantic Bias, Deep Learning},
}

\begin{document}
\maketitle

\begin{abstract}
	Speech is a scalable and non-invasive biomarker for early mental health screening. However, widely used depression datasets like DAIC-WOZ exhibit strong coupling between linguistic sentiment and diagnostic labels, encouraging models to learn semantic shortcuts. As a result, model robustness may be compromised in real-world scenarios, such as Camouflaged Depression, where individuals maintain socially positive or neutral language despite underlying depressive states. To mitigate this semantic bias, we propose DepFlow, a three-stage depression-conditioned text-to-speech framework. First, a Depression Acoustic Encoder learns speaker- and content-invariant depression embeddings through adversarial training, achieving effective disentanglement while preserving depression discriminability (ROC-AUC: 0.693). Second, a flow-matching TTS model with FiLM modulation injects these embeddings into synthesis, enabling control over depressive severity while preserving content and speaker identity. Third, a prototype-based severity mapping mechanism provides smooth and interpretable manipulation across the depression continuum. Using DepFlow, we construct a Camouflage Depression-oriented Augmentation (CDoA) dataset that pairs depressed acoustic patterns with positive/neutral content from a sentiment-stratified text bank, creating acoustic-semantic mismatches underrepresented in natural data. Evaluated across three depression detection architectures, CDoA improves macro-F1 by 9\%, 12\%, and 5\%, respectively, consistently outperforming conventional augmentation strategies in depression Detection. Beyond enhancing robustness, DepFlow provides a controllable synthesis platform for conversational systems and simulation-based evaluation, where real clinical data remains limited by ethical and coverage constraints.
\end{abstract}

% keywords can be removed

\section{Introduction}
\label{sec:introduction}
Depression alters speech production in systematic ways, including reduced vocal energy and imprecise articulation \cite{cummins2015review, low2010detection}. These changes are associated with autonomic dysregulation and reflect core clinical symptoms such as psychomotor retardation and reduced affective expressiveness \cite{bandyopadhyay2014american, mccorry2007physiology, almaghrabi2023bio}. Importantly, such acoustic markers often emerge before individuals consciously recognize or report their own depressive symptoms \cite{morales2025mapping, brown2025camouflaging}. As a result, speech constitutes a scalable, non-invasive, and relatively difficult to intentionally manipulate biomarker for early and passive mental health screening, particularly in real-world monitoring settings where robustness across expression styles is essential.

Speech conveys information through multiple interacting layers, including acoustic cues, linguistic content, and speaker identity \cite{scherer2003vocal}. While all three contribute to affective expression, our analysis identifies linguistic sentiment as a critical confounder in the widely used DAIC-WOZ dataset. Concretely, by examining nearly ten thousand utterances in DAIC-WOZ, we find that negative sentiment is substantially more frequent in depressed than in healthy speech. This dataset-induced coupling indicates that linguistic sentiment is not independent of depression status, meaning that models trained on such data are naturally encouraged to rely on sentiment cues rather than depression-related acoustic markers \cite{geirhos2020shortcut, pearl2009causality, scholkopf2021toward}.

This risk is particularly pronounced when linguistic content and acoustic expression diverge, for example when individuals with depression maintain socially positive or neutral language while still exhibiting depressive vocal patterns. This aligns with the clinical notion of Camouflaged Depression \cite{jansson2016psychiatric, brown2025camouflaging, shetty2018understanding}, where individuals consciously or unconsciously mask symptoms to conform to social expectations. In these cases, acoustic-semantic mismatches are not anomalies but can occur systematically in certain populations. Models that treat linguistic sentiment as a proxy for depression are therefore prone to failure whenever semantic content no longer reflects internal affective state, leading to degraded robustness in deployment.

To mitigate semantic bias and reduce shortcut learning, we propose to explicitly expose models to controlled conflicts between acoustic and semantic information during training. To this end, we introduce DepFlow, a waveform-level, depression-conditioned speech generation framework capable of synthesizing the Camouflage Depression-oriented Augmentation (CDoA) dataset, where depression-related acoustic expressiveness and linguistic content can be manipulated independently while preserving speaker identity. Rather than generating arbitrary synthetic data, we specifically construct datasets that simulate Camouflaged Depression scenarios: for depressed speakers, DepFlow produces additional speech in which their original depressive acoustic patterns are retained but paired with externally provided positive linguistic content, a combination that is severely underrepresented in real datasets; for healthy speakers, additional speech with predominantly positive content is generated as well. By providing such systematically constructed, high-fidelity speech in which depressive acoustics are intentionally placed in conflict with positive semantics, DepFlow encourages downstream models to rely less on semantic polarity and instead shift decision boundaries toward invariant acoustic biomarkers.

DepFlow achieves this through three core components. First, a Depression Acoustic Encoder (DAE) learns a continuous manifold of depression-related acoustic patterns while suppressing speaker identity and linguistic information via adversarial objectives. Second, a conditional flow-matching text-to-speech model synthesizes waveform-level speech conditioned on these depression embeddings, enabling precise control over depressive acoustic expressiveness while preserving linguistic content and speaker traits. Finally, a prototype-based severity control mechanism provides smooth and interpretable manipulation of depressive expressiveness at inference time, allowing DepFlow to systematically generate Camouflage-oriented training samples that break the sentiment–diagnosis correlation present in existing datasets.

We evaluate DepFlow by augmenting the training data of three representative deep learning–based depression detection architectures. Across all models, incorporating DepFlow-generated disentangled speech consistently improves macro-F1 scores, indicating more stable and robust decision boundaries. Moreover, DepFlow consistently outperforms existing data augmentation strategies, highlighting the benefit of targeted acoustic–semantic decoupling over generic perturbation-based approaches. These results demonstrate that exposing models to depressive acoustics paired with non-depressive linguistic content effectively reduces semantic shortcut learning and shifts model reliance toward invariant acoustic biomarkers.

Beyond improving robustness in depression detection, DepFlow also serves as a depression-oriented speech data generation tool that supports emerging research paradigms in mental health modeling. In particular, dialogue-based and interactive studies of depression require large-scale, diverse, and controllable speech data \cite{zhang2023speechgpt, ji2024wavchat}, which are difficult to obtain from real-world clinical recordings due to ethical constraints, privacy concerns, and limited coverage of expressive variability \cite{cummins2015review, li2025automated}. This data-generation capability paves the way for future research on depression-aware conversational systems, interaction modeling, and simulation-based evaluation frameworks, where relying solely on real data is insufficient to support training and analysis.

The remainder of this paper is organized as follows. Section II reviews related work on depression detection and data augmentation. Section III describes the proposed DepFlow framework and provides detailed descriptions of its component modules. Section IV outlines the experimental setup, including datasets, preprocessing, implementation details, construction of CDoA datasets, downstream depression detection architectures, and evaluation metrics. Section V presents the experimental results and discusses the key findings. Section VI concludes the paper. 
\section{Related Work}
\subsection{Speech-based Depression Detection using Deep Learning}

Much of the progress in this field has been driven by the AVEC challenge series \cite{valstar2013avec, ringeval2019avec}, which established standard benchmarks for multimodal depression assessment.

Early systems relied on handcrafted acoustic features such as prosodic cues (pitch, energy, speech rate), spectral descriptors (MFCCs, formants), and voice quality measures (jitter, shimmer, HNR) \cite{li2025automated}. These approaches captured clinically relevant correlates of psychomotor retardation but required manual design and lacked robustness.

End-to-end models addressed these limitations by learning directly from raw audio or mel-spectrograms. CNN-based architectures extract local spectral patterns \cite{chlasta2019automated, dubagunta2019learning, saidi2020hybrid, Othmani2021, vazquez2020automatic, wang2022ecapa}, while RNNs model temporal dynamics and Transformers capture long-range dependencies \cite{Salekin2018, zhao2020hierarchical, muzammel2020audvowelconsnet}.

More recently, self-supervised learning (SSL) has become the dominant paradigm due to its ability to learn rich acoustic representations from large unlabeled corpora. Models such as Wav2Vec 2.0 \cite{baevski2020wav2vec}, HuBERT \cite{hsu2021hubert}, WavLM \cite{chen2022wavlm}, and Whisper \cite{radford2023robust} have shown strong transferability. SSL-based features consistently outperform handcrafted and conventional end-to-end methods, particularly in low-resource depression detection tasks \cite{Zhang2021Depa, Toto2021, chen2022speechformer, ravi2022step, wu2023self, li2025hierarchical,zhang2025speecht}.

Despite these advances, most models implicitly learn the strong correlation between linguistic sentiment and depression labels present in training data. This semantic bias makes models brittle whenever textual sentiment and acoustic affect diverge.

\begin{figure*}[htp]
    \centering
    \includegraphics[width=1.0\linewidth]{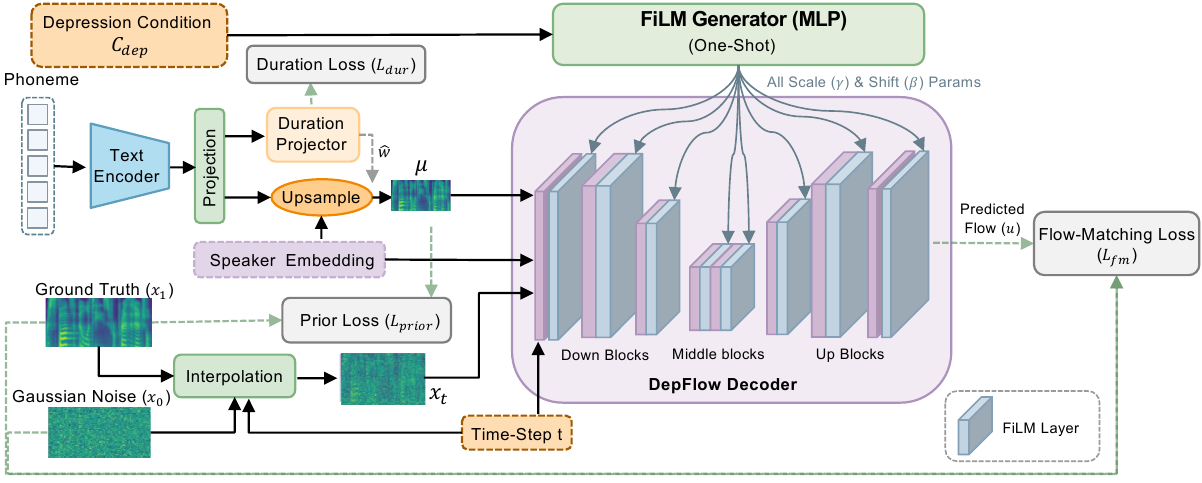}
    \caption{\textbf{Training pipeline of DepFlow}. DepFlow takes phoneme sequences, speaker embeddings, and a depression condition embedding $\mathbf{c}_{\mathrm{dep}}$ as conditioning inputs. A text encoder provides linguistic features and a duration module aligns them with acoustic frames, while a Conditional Flow Matching decoder generates mel-spectrograms with FiLM-based depression conditioning.}
    \label{fig:overall_structure}
\end{figure*}

\subsection{Data Augmentation for Speech-based Depression Detection}

Data augmentation approaches can be grouped into intra-class and inter-class strategies \cite{lishi2025vector}.

\subsubsection{Intra-Class Augmentation}

Intra-class methods perturb samples without altering their diagnostic labels. Traditional variants apply pitch shifts, speed modifications, or additive noise. More advanced techniques such as FrAUG \cite{ravi2022fraug} and SpecAugment \cite{park2019specaugment} operate on time–frequency representations to improve robustness. Domain-aware approaches further manipulate structural or phoneme-level properties to introduce targeted variability, such as dialogue shuffling \cite{wu2023self}, vowel-level perturbation \cite{feng2023knowledge}, or spectrogram-resolution variation \cite{kumar2024depression}.

While effective, these methods only enrich within-class diversity. They do not address the label–feature entanglement across diagnostic categories, nor do they mitigate semantic bias arising from sentiment–diagnosis correlations.

\subsubsection{Inter-Class Augmentation}

Inter-class augmentation aims to generate samples that reflect transformations across diagnostic categories. Mixup-style interpolation has been explored, but its assumption of linear label mixing often fails in clinical tasks, producing pathologically invalid samples \cite{lishi2025vector}. GAN-based approaches \cite{goodfellow2020generative} have attempted class-to-class transformations, including affective style transfer via CycleGAN variants \cite{yang2020feature}. However, GANs struggle to provide stable, fine-grained control over depression severity and often operate in feature space rather than at the waveform level.

Counterfactual augmentation offers a more principled alternative by explicitly modifying latent representations. Zuo and Mak \cite{lishi2025vector} proposed VQ-based counterfactual manipulation to map an utterance toward an opposite diagnostic class, demonstrating improved robustness under data scarcity.

With advances in neural speech synthesis, waveform-level augmentation has gained traction. DepressGEN \cite{liang2025depressgen} uses controllable TTS to generate coherent depression-like samples, combining LLM-driven text generation with prosody control. However, prior systems generally rely on prosodic imitation rather than disentangled acoustic–semantic manipulation, and they do not directly target the correction of semantic bias.

\section{Methodology}

\subsection{Overview}
Figure~\ref{fig:overall_structure} illustrates the overall architecture of our system, which is composed of three components: the Depression Acoustic Encoder (DAE), the DepFlow TTS model, and a prototype-based severity mapping mechanism. The framework separates the learning of depressive acoustics from speech synthesis: the DAE first learns a speaker- and content-invariant manifold of depressive acoustic patterns, and DepFlow then uses this embedding to synthesize disentangled speech, injecting depressive acoustic cues into arbitrary text while preserving speaker identity. Finally, the severity mapping module provides continuous and interpretable control over depressive expressiveness, enabling the system to generate Camouflage-oriented speech samples that systematically break the sentiment–diagnosis correlation present in standard datasets.

\subsection{Depression Acoustic Encoder}
\begin{figure}
    \centering
    \includegraphics[width=0.7\linewidth]{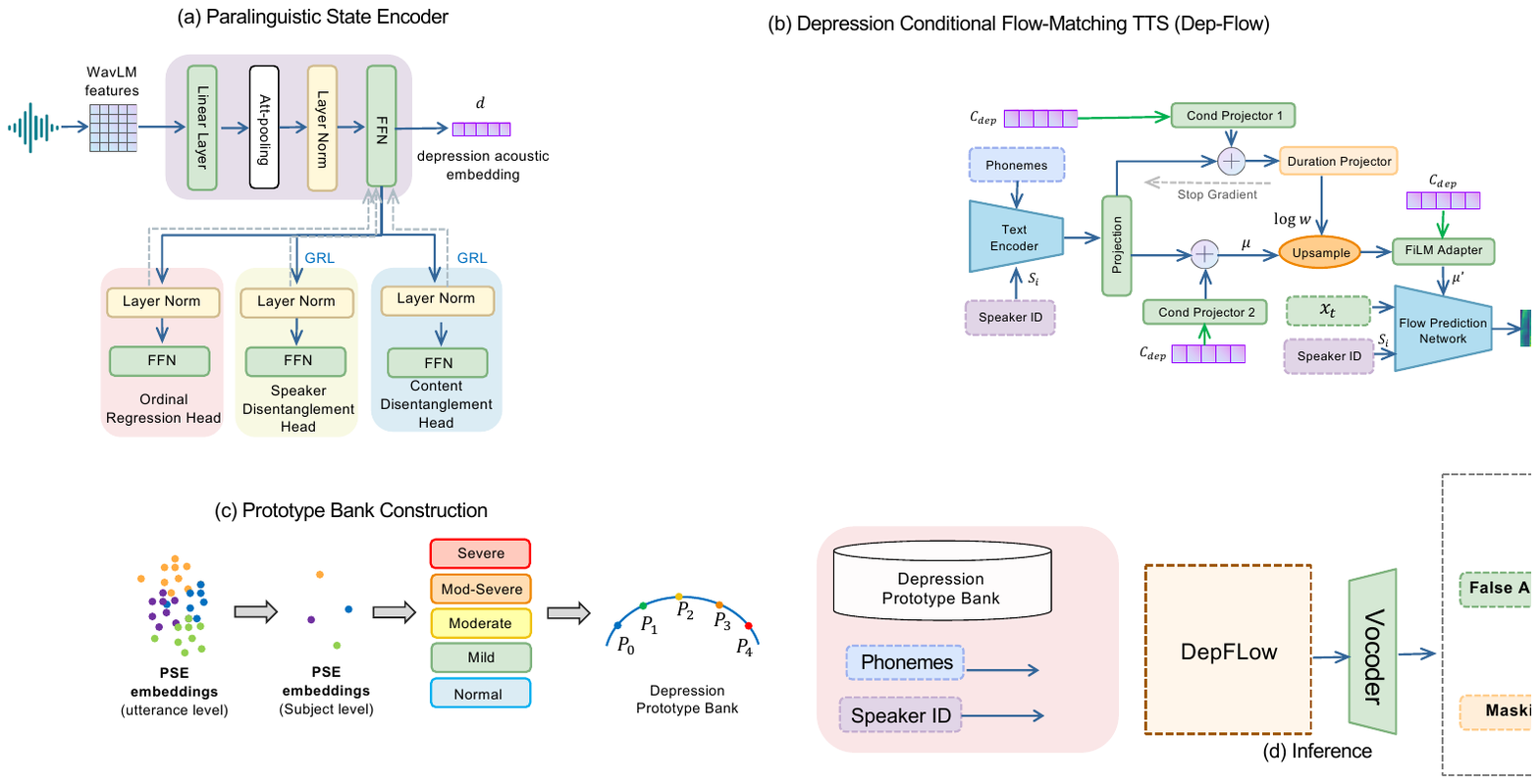}
    \caption{\textbf{Architecture of the Depression Acoustic Encoder (DAE).} Frame-level WavLM features are aggregated into an utterance-level representation to produce a depression acoustic embedding $\mathbf{d}$, with an ordinal regression head estimating PHQ-based severity and GRL-based speaker and content disentanglement heads suppressing speaker identity and linguistic information.}
    \label{fig:DAE}
\end{figure}

To separate depression-related acoustics from linguistic and speaker factors, we design a Depression Acoustic Encoder (DAE) that learns a continuous manifold of depressive acoustic patterns while abstracting away speaker identity and linguistic content. This disentanglement is essential for disentangled speech generation: depressive cues must be modified without distorting phonemes or altering speaker traits, otherwise the synthesized samples would be clinically unreliable for decision-boundary correction. The architecture of the DAE is shown in Figure \ref{fig:DAE}.

\subsubsection{Model Structure}
A pretrained WavLM-Large model~\cite{chen2022wavlm} first extracts frame-level contextualized features $\{\mathbf{x}_t\}_{t=1}^{T}$. Each frame is projected to a 256-dimensional embedding through a linear layer with ReLU and dropout, producing $\mathbf{h}_t$. An attention-based statistical pooling layer then aggregates these frame embeddings into an utterance-level representation by computing attention weights $\alpha_t=\mathrm{softmax}(f{\text{attn}}(\mathbf{h}_t))$ and concatenating the attention-weighted mean and standard deviation.

A post-encoder composed of fully connected layers with Layer Normalization, SiLU activation, and dropout transforms this pooled embedding into a 32-dimensional depression-related acoustic embedding $\mathbf{d}$, which serves as the shared representation for all downstream branches.

On top of this embedding, three lightweight heads are attached:

\paragraph{Ordinal Regression Head}
This head predicts $K\!-\!1=4$ ordered thresholds for the $K=5$ PHQ-8 severity levels ~\cite{kroenke2009phq}. Given $\mathbf{d}$, a small MLP outputs threshold logits.

\paragraph{Speaker Disentanglement Head}  
To enforce speaker invariance, a speaker classifier is applied to the L2-normalized embedding $\tilde{\mathbf{d}}$ and predicts over $S$ speakers. The same classifier is reused in an adversarial branch equipped with a gradient reversal layer (GRL)~\cite{ganin2016domain}, which reverses gradients to discourage speaker-specific information in the shared embedding.

\paragraph{Linguistic Disentanglement Head}
A linguistic-adversarial branch reduces residual phonetic content in $\tilde{\mathbf{d}}$. A pretrained HuBERT model~\cite{hsu2021hubert} generates frame-level pseudo-phoneme labels $\{c_t\}$, which are collapsed to an utterance-level category via majority voting. The normalized embedding is passed through another GRL and a classifier $f_{\text{con}}$ predicting over $C$ HuBERT-derived units. This adversarial signal suppresses linguistic information.

\subsubsection{Optimization Objectives}
Training jointly optimizes depression prediction and the removal of speaker and linguistic information from the shared embedding $\mathbf{d}$. The Ordinal Regression Head is trained using binary cross-entropy on monotonic ordinal targets:
\begin{equation}
\mathcal{L}_{\text{sup}}
= -\sum_{k=1}^{K-1}
\Bigl[t_k \log \sigma(o_k) + (1 - t_k)\log \bigl(1 - \sigma(o_k)\bigr)\Bigr],
\end{equation}
with class-balanced weights derived from label frequencies.

The speaker classifier applied to $\tilde{\mathbf{d}}$ uses a cross-entropy identification loss $\mathcal{L}_{\text{id}}$, while its adversarial branch contributes an adversarial speaker loss $\mathcal{L}_{\text{adv-spk}}$ encouraging speaker-invariant embeddings. Similarly, the linguistic-adversarial branch is trained with a cross-entropy loss $\mathcal{L}_{\text{adv-con}}$ over HuBERT-derived labels, promoting suppression of linguistic content.

The full training objective is a weighted sum:
\begin{equation}
\mathcal{L}_{\text{total}}
= \lambda_{\text{sup}}\mathcal{L}_{\text{sup}}
+ \lambda_{\text{id}}\mathcal{L}_{\text{id}}
+ \lambda_{\text{spk}}\mathcal{L}_{\text{adv-spk}}
+ \lambda_{\text{con}}\mathcal{L}_{\text{adv-con}}.
\end{equation}

\begin{figure*}
    \centering
    \includegraphics[width=1\linewidth]{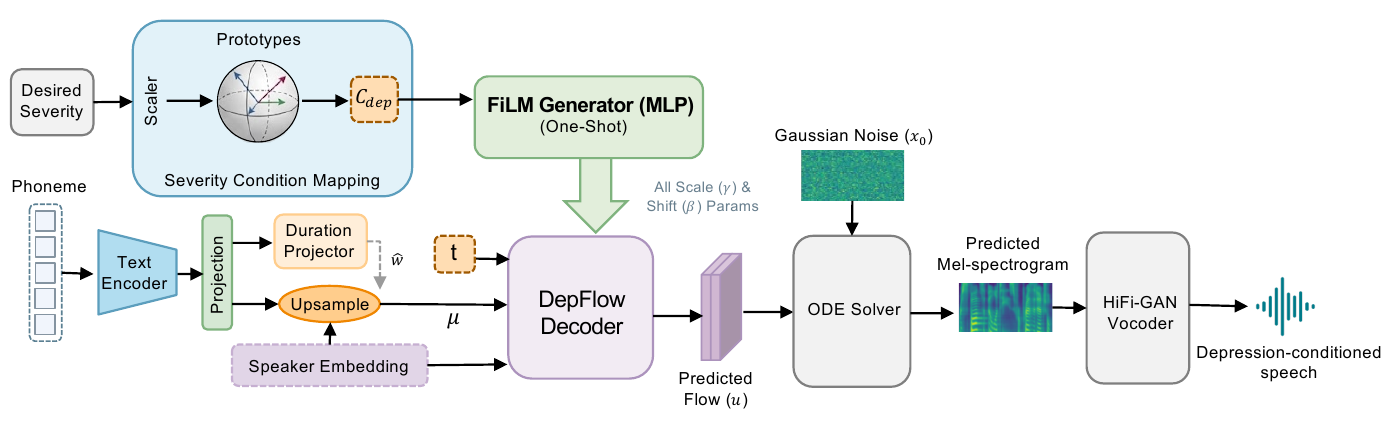}
    \caption{\textbf{Inference pipeline of DepFlow.} Given a desired severity level, DepFlow maps it to a depression condition embedding $\mathbf{c}_{\mathrm{dep}}$ and, together with phoneme and speaker inputs, generates severity-controlled mel-spectrograms via FiLM-conditioned decoding, which are finally converted to speech using HiFi-GAN.}
    \label{fig:Inference}
\end{figure*}
\subsection{Depression Conditioned Flow Matching TTS (DepFlow)}

DepFlow adapts the Matcha-TTS architecture \cite{mehta2024matcha} to enable waveform-level synthesis conditioned on the DAE-derived depression embedding. The model comprises a text encoder, a duration predictor, and a Conditional Flow Matching (CFM) decoder, designed to satisfy three key requirements: high-fidelity synthesis, global coherence, and precise controllability \cite{lipman2022flow,zhang2024speaking}.

During synthesis, the CFM decoder receives three inputs, namely a phoneme sequence (text), a speaker embedding and a DAE-derived depression embedding. The depression embedding does not modify linguistic or speaker representations; instead, it acts as an external style variable that shapes vocal energy, prosody, and articulation.

\subsubsection{FiLM-based Depression Conditioning}

A key requirement for disentangled speech generation is global consistency: depressive cues must affect all acoustic scales, from frame-level detail to long-range prosody. To meet this requirement, DepFlow adopts a FiLM-based modulation mechanism \cite{perez2018film} instead of feature concatenation or attention mechanisms.

Concatenation provides weak conditioning and often fails to influence deeper decoder layers, while cross-attention introduces unnecessary computational overhead. FiLM, in contrast, applies channel-wise affine transformations to intermediate activations, offering both efficiency and uniform influence across the entire decoding hierarchy.

Given a depression control embedding $\mathbf{c}_{\text{dep}}$, a lightweight MLP produces scale–shift pairs $(\gamma_i, \beta_i)$ for all FiLM-modulated blocks. Each decoder block applies channel-wise affine modulation:
\begin{equation}
\hat{\mathbf{h}}_i = \gamma_i \cdot \mathbf{h}_i + \beta_i .
\end{equation}
where $\mathbf{h}\in\mathbb{R}^{B\times C\times T}$ denotes the intermediate feature map.

FiLM acts as a global conditioning mechanism that scales and shifts the feature maps, effectively modulating the entire energy and pacing of the speech without altering the underlying linguistic structure encoded in the content features.

\subsubsection{DepFlow Decoder}

The DepFlow decoder follows the U-Net architecture of Matcha-TTS \cite{mehta2024matcha}, with downsampling, bottleneck, and upsampling stages. Each stage integrates ResNet and lightweight transformer layers to capture both short-term and long-term acoustic dependencies. Flow matching trains the decoder to predict velocity fields that transport noise into mel-spectrograms, yielding a stable and interpretable continuous-time generative process. Unlike standard diffusion models that rely on complex curved denoising trajectories and may suffer from discretization sensitivity, Flow Matching with Optimal Transport learns smooth and deterministic transport paths. This geometric simplicity is particularly important for our disentangled severity control objective: it ensures that continuous adjustments of the depression-severity condition result in smooth and well-behaved shifts in the generated acoustic trajectory, maintaining stability and speaker identity while avoiding artifacts commonly introduced by nonlinear interpolation.

\subsubsection{Optimization Objective}

Training uses duration loss, Gaussian prior loss, and flow matching loss as in Matcha-TTS:

\paragraph{Duration Loss}
\begin{equation}
\mathcal{L}_{\text{dur}} = \text{MSE}(\log \mathbf{w}, \log \hat{\mathbf{w}}).
\end{equation}
where $\mathbf{w}$ and $\hat{\mathbf{w}}$ denote ground-truth and predicted durations.

\paragraph{Prior Loss}
A Gaussian prior encourages consistency between encoder outputs and target mel-spectrograms:
\begin{equation}
\mathcal{L}_{\text{prior}} = \frac{1}{|\mathcal{M}| \cdot n_f} 
\sum_{(i,j) \in \mathcal{M}}
\frac{1}{2}
\Big[(y_{i,j} - \mu_{i,j})^2 + \log(2\pi)\Big].
\end{equation}

\paragraph{Flow Matching Loss}
Following the CFM formulation, we train the decoder to predict the velocity field:
\begin{equation}
\mathcal{L}_{\text{fm}} = 
\frac{1}{|\mathcal{M}| \cdot n_f}
\sum_{(i,j) \in \mathcal{M}}
\left\|
\mathbf{u}_{i,j} - \hat{\mathbf{u}}_{i,j}
\right\|^2 .
\end{equation}
where the target velocity is
\begin{equation}
\mathbf{u} = \mathbf{x}_1 - (1 - \sigma_{\min})\mathbf{z}.
\end{equation}

\paragraph{Total Loss}
\begin{equation}
\mathcal{L}_{\text{total}} 
= \mathcal{L}_{\text{dur}} 
+ \lambda_p \mathcal{L}_{\text{prior}} 
+ \mathcal{L}_{\text{fm}} .
\end{equation}
where $\lambda_p$ is set to 1.0.

\subsection{Prototype-based Severity Condition Mapping}
Although DepFlow enables depression-conditioned generation, effective disentangled speech synthesis requires a severity representation that is both clinically meaningful and smoothly controllable. Raw DAE embeddings, while informative, may exhibit inter-speaker variability or nonlinear geometry that complicates direct manipulation. To obtain a stable and interpretable severity axis, we construct a prototype bank representing the central depressive acoustic pattern at each PHQ-8 severity level. During synthesis, these prototype-conditioned severity embeddings serve as the depression control variable fed into DepFlow, enabling fine-grained manipulation of depressive acoustic expressiveness.

\subsubsection{Subject-level Embedding Aggregation}

Depression severity manifests more reliably across a subject’s recordings than within individual utterances. We therefore compute a subject-level embedding $\mathbf{d}^{(\text{subj})}$ by averaging utterance-level DAE embeddings. This reduces within-speaker variance and yields a robust depression descriptor aligned with clinical assessments.

Subjects are grouped into the five PHQ-8 severity bins used during DAE training. For each bin $k$, we compute a prototype:
\begin{equation}
\bar{\mathbf{p}}_k = \frac{1}{N_k} \sum_{j \in \mathcal{S}_k} \mathbf{d}^{(\text{subj})}_j,\quad
\mathbf{p}_k = \frac{\bar{\mathbf{p}}_k}{\|\bar{\mathbf{p}}_k\|_2},
\end{equation}
The prototype $\mathbf{p}_k$ represents the central depressive acoustic pattern of each severity category, abstracted from individual idiosyncrasies.

\subsubsection{Continuous Severity Control via SLERP}

During inference, the goal is to generate a depression embedding corresponding to any desired PHQ-8 score, not just the five discrete bins. Given a target score $s$, we map it to a normalized scalar: 

\begin{equation}
\alpha(s) = \mathrm{clip}\!\left(\frac{s - 12}{12},\, -1,\, 1\right).
\end{equation}

which specifies its relative position in the severity spectrum. We then identify the two adjacent prototypes $\mathbf{p}_i$ ,$\mathbf{p}_{i+1}$ and compute an interpolation weight $\tau(s)\in[0,1]$. The final severity embedding is obtained using spherical linear interpolation (SLERP) \cite{shoemake1985animating}: 

\begin{equation}
\mathbf{c}_{\mathrm{dep}}^{\mathrm{infer}}(s) = \mathrm{slerp}\bigl(\mathbf{p}_i, \mathbf{p}_{i+1};\, \tau(s)\bigr).
\end{equation}

SLERP preserves angular geometry on the unit hypersphere, producing smooth transitions in depressive expressiveness while avoiding artifacts associated with linear interpolation. 

\section{Experimental Setup}
\begin{table*}[htbp]
\caption{\textbf{Demographic and diagnostic distribution of the DAIC-WOZ dataset.}
\label{tab:DAICWOZ}}
\centering
\begin{tabular}{cccccccc}
\toprule
\multirow{2}{*}{Gender} & \multirow{2}{*}{Category} & \multirow{2}{*}{Number} & \multicolumn{4}{c}{PHQ-8 Score} & \multirow{2}{*}{PHQ Score Statistics} \\

 & & & 0-4 & 5-9 & 10-19 & 20-24 & \\
\midrule
Female & Control Group & 56 &38 & 18& 0& 0& 3.4±3.09 \\

Female & Depression Group & 31 & 0&0 & 25& 6& 14.5±4.19 \\

Male & Control Group & 76 & 48& 28& 0& 0& 6.8±5.92 \\

Male & Depression Group & 26 & 0& 0& 25& 1& 14.0±3.22 \\
\bottomrule
\end{tabular}

\end{table*}
\begin{figure*}
    \centering
    \includegraphics[width=0.8\linewidth]{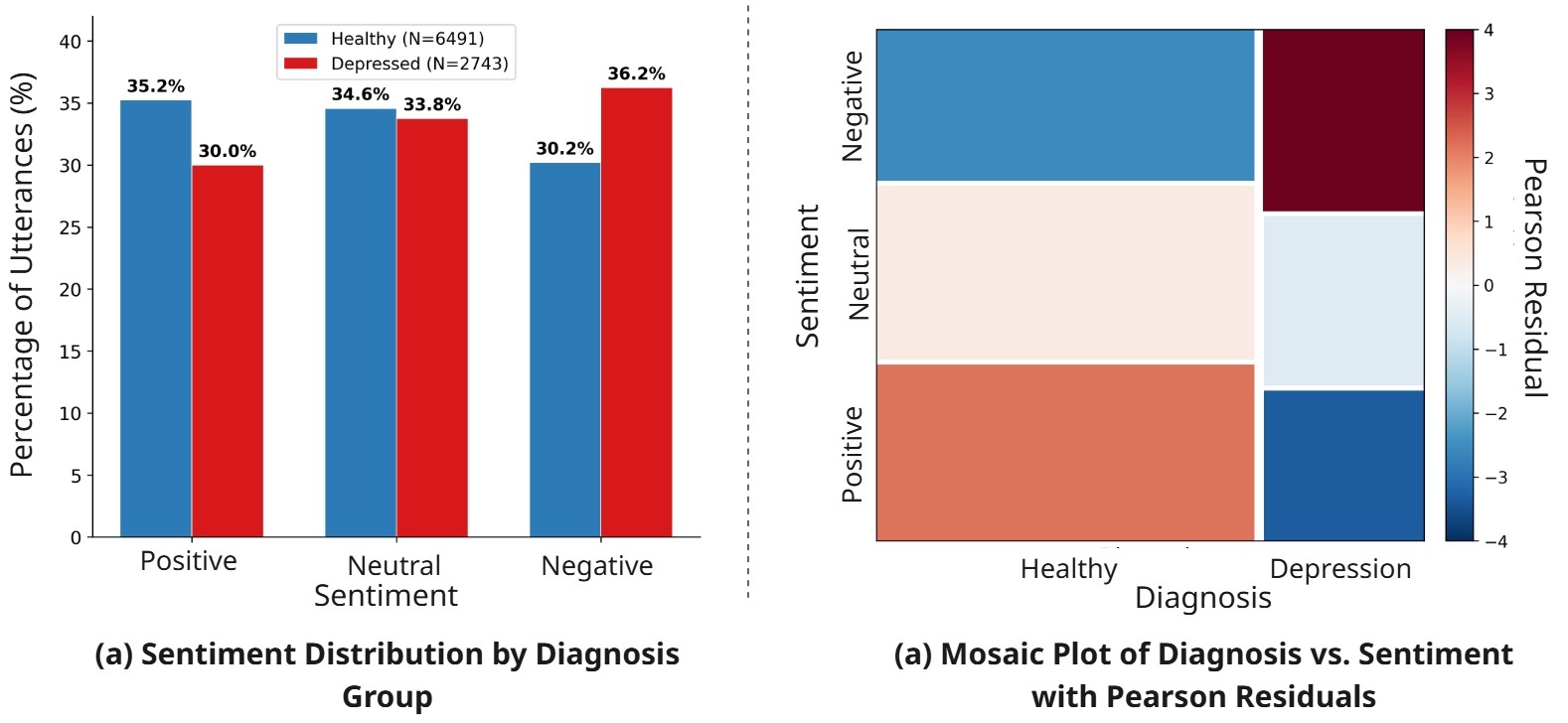}
    \caption{Analysis of semantic bias in the DAIC-WOZ dataset. \textbf{(a)} Sentiment Distribution by Diagnosis Groups; \textbf{(b)} Mosaic plot of Diagnosis vs.\ Sentiment shaded by Pearson residuals. Red tiles indicate overrepresentation (observed $>$ expected), while blue tiles indicate underrepresentation. The deep red shading in the \textit{Depressed-Negative} intersection highlights a significant correlation between depression labels and negative linguistic sentiment.}
    \label{fig:sentiment_dist}
\end{figure*}

\subsection{Datasets}
Two datasets are used in our experiments: the CSTR VCTK corpus~\cite{yamagishi2019cstr} and the DAIC-WOZ dataset~\cite{gratch2014distress}. The CSTR VCTK corpus contains recordings from 110 English speakers, originally sampled at 48~kHz. The DAIC-WOZ dataset consists of clinical interviews from 189 English-speaking participants, including 56 clinically diagnosed with depression, with audio sampled at 16~kHz. Notably, 409 samples in the DAIC-WOZ training partition are relabeled from 0 to 1 following the correction reported in~\cite{bailey2021gender}\footnote{https://github.com/adbailey1/daic\_woz\_process}.

In our pipeline, CSTR VCTK is used exclusively to pretrain the text-to-speech backbone (DepFlow base model), while DAIC-WOZ is used to (i) train the Depression Acoustic Encoder (DAE), (ii) finetune DepFlow with depression conditioning, and (iii) train and evaluate downstream depression detection models.

All operations that require depression labels strictly use only the DAIC-WOZ training and development sets, and the official test set is reserved for final evaluation only.

\subsection{Sentiment Distribution Analysis}

We analyze the relationship between linguistic sentiment and depression diagnosis in the DAIC-WOZ training set. Sentiment analysis was applied to 9,234 interviewee utterances, diarized by timestamps and excluding segments shorter than 1~s. Each utterance was classified as positive, neutral, or negative using the DeepSeek R1 large language model~\cite{guo2025deepseek} \footnote{https://github.com/deepseek-ai/DeepSeek-R1}. Sentiment distributions were computed separately for healthy and depressed participants.

Figure~\ref{fig:sentiment_dist}(a) shows a clear distributional shift between groups. Healthy participants produced more positive than negative utterances (35.2\% vs. 30.2\%), whereas depressed participants exhibited the opposite pattern, with negative sentiment exceeding positive sentiment (36.2\% vs. 30.0\%).

To quantify this association, we visualized the sentiment-diagnosis contingency table using a mosaic plot shaded by Pearson residuals (Figure~\ref{fig:sentiment_dist}(b)). For each cell $(i,j)$, the Pearson residual is defined as
\begin{equation}
    r_{ij} = (O_{ij} - E_{ij}) / \sqrt{E_{ij}},
\end{equation}
where $O_{ij}$ and $E_{ij}$ denote observed and expected frequencies under independence. These patterns are statistically significant, as confirmed by Pearson residuals and a chi-square test of independence.

This association is further confirmed by a Pearson chi-square test of independence ($\chi^2 = 38.03$, $df = 2$, $p < 10^{-8}$)~\cite{pearson1900x}. Overall, linguistic sentiment in DAIC-WOZ is strongly coupled with diagnostic labels, suggesting the presence of a semantic shortcut exploitable by classification models.

\subsection{Preprocessing and Data Splits}
For all experiments, we adopt the official DAIC-WOZ partition (train: 107 subjects, development: 35 subjects, test: 47 subjects), ensuring subject-level disjointness across splits.

For both DAE training and downstream depression detection experiments, we follow the Speechformer~\cite{chen2022speechformer} sampling protocol for class balancing at the utterance level. During training, a 10-second segment is randomly cropped from each selected utterance in every epoch to introduce temporal variability. For evaluation, the 20 longest utterances from each test subject are used, and subject-level predictions are obtained by majority voting over utterance-level outputs.

For DepFlow pretraining on VCTK, we use the original corpus segmentation provided by the dataset and downsample audio from 48~kHz to 22.05~kHz to match our vocoder and acoustic feature pipeline.  

For DepFlow finetuning on DAIC-WOZ, all utterances from the training and development sets are used after removing segments shorter than one second or dominated by silence. These recordings are resampled from 16~kHz to 22.05~kHz. No information from the DAIC-WOZ test set is used during training or generation.

All DepFlow-generated speech samples are produced strictly from the DAIC-WOZ training set and are only used to augment the training data. They are never introduced into the development or test sets.

\subsection{Implementation Details}

All models are trained on NVIDIA L40 GPUs. The training of the generation framework consists of three stages.

\paragraph{Stage 1: Depression Acoustic Encoder}
The DAE is optimized for up to 500 epochs with a batch size of 64 using AdamW~\cite{loshchilov2017decoupled} (learning rate $1\times10^{-4}$, weight decay $3\times10^{-3}$). A dropout rate of 0.2 is applied to all projection and prediction layers. The loss weights are set to $\lambda_{\text{sup}}=1.0$, $\lambda_{\text{id}}=0.2$, $\lambda_{\text{spk}}=0.2$, and $\lambda_{\text{con}}=0.1$. Early stopping is applied based on performance on the DAIC-WOZ development set.

\paragraph{Stage 2 \& 3: Pretraining and Finetuning DepFlow}
We first pretrain the base Matcha-TTS model on VCTK without depression conditioning, following the default parameter settings in Matcha-TTS \cite{mehta2024matcha} unless otherwise specified. DepFlow is then finetuned on DAIC-WOZ with depression condition and FiLM generator, using the same hyperparameter settings. The depression conditioning embedding has a dimensionality of 32. During finetuning, both the base TTS parameters and the FiLM generator are updated. The dropout rate of the FiLM generator is set to 0.2. During inference, waveforms are reconstructed from mel-spectrograms using a pretrained HiFi-GAN vocoder \cite{kong2020hifi}.

\subsection{Evaluation Protocols}
\subsubsection{Controllability and Validity of Synthesized Speech}
\label{sec:severity_protocol}

A primary objective of DepFlow is to modulate depressive acoustic expressiveness in a controlled manner while keeping linguistic content and speaker identity fixed. We therefore evaluate controllability and validity from three complementary perspectives. First, we assess whether increasing the intended severity produces monotonic shifts within the learned depression embedding manifold. Second, we examine whether these latent shifts translate into systematic changes in clinically relevant acoustic markers at the intra-speaker level. Third, we verify that severity manipulation does not compromise TTS quality, including linguistic intelligibility and speaker similarity.

All analyses in this section are conducted under a unified severity-controlled synthesis protocol applied to the DAIC-WOZ test set. For each utterance, five synthetic variants are generated to correspond to the five clinical severity levels from healthy to severe depression. For each target level, the PHQ-derived severity value is converted into a continuous scalar $\alpha !\in! [-1,1]$ and subsequently mapped to a depression condition embedding $\mathbf{c}_{\mathrm{dep}}$, which, together with the target speaker identity, conditions DepFlow for generation.

This protocol yields five severity-controlled speech versions per utterance, resulting in 4,700 synthesized test samples. It is used strictly for evaluation; no synthesized samples are included in model training, thereby preventing any form of data leakage into downstream depression detection experiments.

\begin{figure*}[htp]
  \centering
  \includegraphics[width=\linewidth]{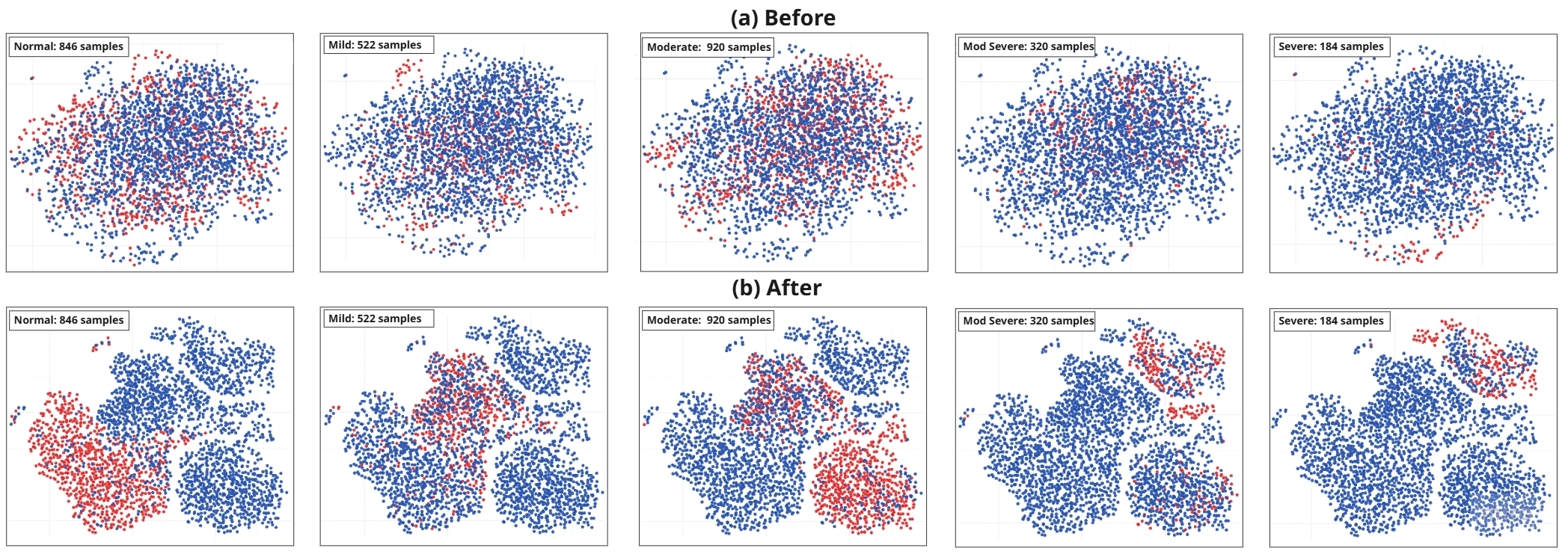}
  \caption{UMAP visualization of the DAE embeddings. (a) Embeddings obtained from a frozen WavLM-Large model; 
(b) Embeddings of the same utterances after processing through the trained DAE. 
In each panel, the five subplots correspond respectively to the five clinical severity levels (Healthy, Mild, Moderate, Moderately Severe, Severe): red points denote samples from the target severity level, while blue points denote all remaining samples.
}
  \label{fig:umap_DAE}
\end{figure*}

\subsubsection{Construction of Camouflage Depression-oriented Augmentation (CDoA) Datasets}
\label{sec:data_construction}
To evaluate the impact of incongruent semantic-acoustic cues, we generate CDoA datasets through a structured three-step procedure.

First, we build a sentiment-based text bank. All DAIC-WOZ training transcripts are annotated at the utterance level (positive, neutral, negative) using a large language model, DeepSeek R1 \cite{guo2025deepseek}. Positive and neutral utterances are grouped into a benign-text bank, while negative utterances form a depressive-text bank, enabling explicit control over semantic polarity.

Following the baseline sampling strategy, we determined synthesis quotas to enforce global binary class balance and coverage across all PHQ severity levels. The final dataset combines original and synthetic data for a total of 5,760 utterances, achieving a perfectly balanced binary distribution (2,880 depressed vs. 2,880 non-depressed). This was achieved by synthesizing a fixed number of utterances per subject for each severity group: 13 (Normal), 34 (Mild), 2 (Moderate), 91 (Moderately Severe), and 194 (Severe).

Finally, we generate disentangled samples by mapping each subject’s PHQ score to a continuous scalar $\alpha \in [-1,1]$ and projecting it into a depression condition embedding $\mathbf{c}_{\mathrm{dep}}$. This embedding conditions DepFlow to inject depressive acoustics into benign (positive or neutral) text, creating controlled acoustic–semantic mismatches where depressive vocal patterns coexist with non-depressive linguistic content.

\paragraph{Baseline Depression Detection Models}
The proposed augmentation method are benchmarked on three representative families of speech-based depression detection models, differing in input features and backbone architectures:

\begin{itemize}
    \item \textbf{DepAudioNet}~\cite{ma2016depaudionet}: A log-Mel-based convolutional–recurrent model. DepAudioNet takes log-Mel spectrograms as input and uses several convolutional layers followed by LSTM layers to encode temporal dynamics, with a fully connected layer for utterance-level depression classification.

    \item \textbf{NUSD}~\cite{wang2023non}: A speaker-disentangling TDNN model operating on frame-level acoustic features. NUSD takes log-Mel and pitch features as input to an ECAPA-TDNN~\cite{desplanques2020ecapa} backbone, and attaches non-uniform speaker-disentangling branches while a main pooling and classification head predicts utterance-level depression labels.

    \item \textbf{HAREN-CTC}~\cite{li2025hierarchical}: A self-supervised hierarchical model built on raw waveforms. HAREN-CTC operates on multi-layer WavLM representations extracted from the raw audio, and employs a hierarchical encoder with a CTC branch and a pooling-based classification head for subject-level depression detection.
\end{itemize}

For all three baselines, we follow the original implementations and hyperparameter settings as closely as possible, and apply our utterance selection and augmentation protocol identically for a fair comparison.

\subsection{Metrics}
The proposed framework are evaluated from four perspectives: (i) disentanglement effect of the learned depression condition embedding from DAE, (ii) controllability of synthesized depression severity, (iii) DepFlow text-to-speech (TTS) quality, (iv) Synthesized CDoA dataset for downstream depression detection performance.

\paragraph{Speaker and Content Disentanglement Metrics}
To verify that the depression embedding learned by the DAE captures severity-relevant structure while suppressing nuisance attributes, both speaker- and content-oriented metrics are employed.

Speaker identity retention is assessed using the Equal Error Rate (EER)~\cite{hansen2015speaker}, a standard speaker verification metric:
\begin{equation}
    \mathrm{EER} = \frac{\mathrm{FAR}(\tau^\ast) + \mathrm{FRR}(\tau^\ast)}{2},
\end{equation}
where $\mathrm{FAR}$ and $\mathrm{FRR}$ denote the False Acceptance Rate and False Rejection Rate, respectively, and $\tau^\ast$ represents the decision threshold where they intersect. Additionally, the \textit{Similarity Gap} is computed, defined as the difference in average cosine similarity between same-speaker and different-speaker pairs.

To evaluate residual linguistic information, a linear probe is trained to regress pooled content representations (HuBERT features~\cite{hsu2021hubert}) from the depression embedding. We report the Mean Squared Error (MSE)~\cite{draper1998applied}, Coefficient of Determination ($R^{2}$)~\cite{draper1998applied}, and Centered Kernel Alignment (CKA)~\cite{kornblith2019similarity} between predicted and ground-truth features.

\paragraph{DepFlow TTS Quality}
Linguistic intelligibility is measured using Word Error Rate (WER), and speaker consistency is assessed using the speaker-similarity metric SIM-o~\cite{le2023voicebox}. WER is computed using Whisper-Large-v3~\cite{radford2023robust}, following the evaluation protocol in~\cite{anastassiou2024seed}. SIM-o is obtained by extracting WavLM-Large embeddings~\cite{chen2022large} and computing cosine similarity between synthesized and corresponding natural utterances.

\paragraph{Severity Controllability in the Latent Space}
To evaluate whether increasing intended severity produces monotonic and globally consistent shifts in the learned depression manifold, we adopt ranking-based metrics. For each utterance group synthesized at the five clinical severity levels, we compute: (i) the Concordance Index (C-index)~\cite{harrell1982evaluating}, which measures the probability that a sample with higher intended severity receives a higher embedding projection value and (ii) Spearman's rank correlation coefficient ($\rho$)~\cite{spearman1987proof}, which evaluates ordinal agreement between severity levels and embedding projections.  
Both metrics emphasize monotonic ordering rather than absolute regression accuracy, making them suitable for validating continuous severity control.

\paragraph{Augmentation Dataset for Depression Detection Performance}
For downstream depression detection models trained with and without DepFlow augmentation, we report subject-level Sensitivity, Specificity, and Macro-F1 in clinical setting ~\cite{sokolova2009systematic}. Subject-level predictions are obtained by majority voting over utterance-level decisions.

\section{Experimental Results and Discussion}
This section evaluates the proposed framework from four complementary perspectives: (1) the quality and structure of the learned depression manifold, (2) the controllability and acoustic validity of DepFlow-synthesized speech, (3) the effectiveness of DepFlow-generated incongruent samples as data augmentation for downstream detectors and (4) qualitative analyses examining how controlled severity manipulation shapes acoustic trajectories.

\begin{figure}
    \centering
    \includegraphics[width=0.5\linewidth]{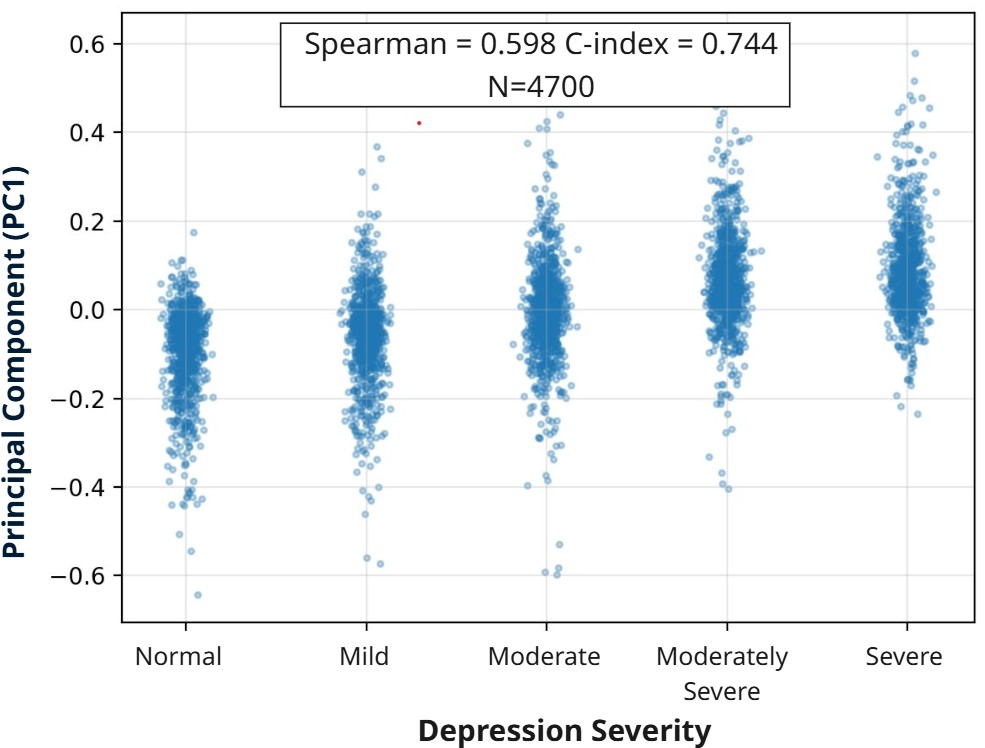}
    \caption{Alignment Between Intended Severity Levels and Latent Depression Embedding Trajectories.}
    \label{fig:emb_contrl}
\end{figure}

\begin{table}
	\caption{Intra-Speaker Acoustic Controllability of DepFlow-Synthesized Speech (N = 4,700 samples).}
	\centering
	\begin{tabular}{ccc}
		\toprule

		\textbf{Feature} & \textbf{Mean $\rho$} & \textbf{Median $\rho$}) \\
		\midrule
        Formant F1 mean            & 0.635 & 0.800 \\
        Formant F2 mean            & 0.604 & 0.800 \\
        Silence–speech ratio      & 0.579 & 0.866 \\
        Formant centralization     & -0.489 & -0.600 \\
        Shimmer                    & 0.304 & 0.500 \\
        HNR                        & -0.259 & -0.500 \\
        Jitter                     & -0.068 & -0.100 \\
        $F_0$ mean                 & 0.064 & 0.100 \\
        $F_0$ std                  & 0.058 & 0.100 \\
		\bottomrule
	\end{tabular}
	\label{tab:table}
\end{table}

\subsection{Structure and Ordinality of the Depression Condition Embedding}

We evaluate whether the proposed Depression Acoustic Encoder (DAE) learns structured and clinically meaningful depressive representations. As illustrated in Fig.~\ref{fig:umap_DAE}, the embedding space exhibits a relatively compact and coherent continuum that shows a gradual transition from healthy to more severe depressive states, whereas frozen WavLM embeddings display weaker clustering and limited ordinal organization. These observations suggest that the DAE tends to organize speech signals along a severity-oriented manifold rather than merely grouping samples into discrete diagnostic categories.

Table III confirms these qualitative observations. The high speaker verification error rate (EER) and low linguistic correlation scores (low $R^2$ and CKA) indicate successful disentanglement of nuisance attributes. Crucially, the embedding retains diagnostic utility, achieving a robust ROC-AUC of 0.693 in linear probing, proving that severity information is preserved even as speaker and content are suppressed.

Overall, these results indicate that the DAE constructs an ordinal depression manifold while empirically weakening speaker- and content-related components, thereby providing a practically useful and clinically relevant conditioning space for downstream disentangled speech generation.

\begin{table}[t]
    \centering
    \caption{Evaluation of disentanglement properties of the proposed DAE embedding on DAIC-WOZ.}
    \begin{tabular}{ccc}
        \toprule
        \textbf{Type} &\textbf{Metrics} & \textbf{Result} \\
        \midrule
        \multirow{2}{*}{Speaker disentanglement} &EER & 0.355 \\
        & Similarity gap (positive-negative) & 0.27 \\
        \midrule
        \multirow{3}{*}{Content disentanglement} &  MSE & 2.83 \\
        & $R^{2}$ & 0.21 \\
        & CKA similarity & 0.014 \\
        \midrule
        \multirow{3}{*}{Depression Classification} & Accuracy & 0.638 \\
        & Macro-F1 & 0.591 \\
        & ROC-AUC & 0.693 \\
        \bottomrule
    \end{tabular}
    \label{tab:dae_disentangle}
\end{table}

\subsection{Controllability and Validity of Synthesized Speech}
All analyses in this section are conducted on the severity-controlled synthetic test set generated using the protocol described in Section~\ref{sec:severity_protocol}.

\subsubsection{Embedding–Space Controllability}

We next examine whether DepFlow produces systematic and ordered movements in the learned depression embedding space as severity increases. For each synthesized utterance, we extract a depression condition embedding $\mathbf{z}_{u,s}$ from pretrained DAE and apply Principal Component Analysis (PCA) to all synthesized embeddings. The first principal component is then used as a shared severity coordinate, and monotonicity with respect to the intended five clinical severity levels is evaluated using the Concordance Index and Spearman's rank correlation.

As shown in Fig.~\ref{fig:emb_contrl}, the distribution of projected embeddings shifts progressively toward higher values from healthy to severe depression. Quantitatively, we obtain a Concordance Index of 0.744 and a Spearman correlation of 0.598, indicating a strong and consistent ordinal relationship between intended severity and the latent trajectory of synthesized speech across speakers and linguistic content. This suggests that DepFlow does not introduce arbitrary variability, but instead drives generation along a coherent direction within the depression manifold learned by the DAE.

\begin{table*}[ht]
\caption{Subject-level depression detection performance under different data augmentation strategies on the DAIC-WOZ dataset. Results are reported as Macro-F1 / Sensitivity / Specificity. Bold denotes the best result for each model.}
\centering
\begin{tabular}{ccccc}
\toprule
Model & Augmentation Methods & Macro-F1 & Sensitivity & Specificity \\
\midrule
\multirow{5}{*}{DepAudioNet~\cite{ma2016depaudionet}} 
 & / & 0.482 (0.042) & 0.346 (0.252) & 0.667 (0.194) \\
 & FrAUG & 0.518 (0.047) &  0.629 (0.137) &  0.495 (0.593) \\
 & Mixup & 0.489 (0.056) &  \textbf{0.657 (0.078)} &  0.418 (0.110) \\
 & Instruct-TTS (CDoA) & 0.459 (0.064)& 0.457 (0.229)&0.539 (0.258)\\
 & DepFlow (CDoA) (Ours) & \textbf{0.526 (0.023)} & 0.417 (0.157) & \textbf{0.674 (0.159)} \\
\midrule
\multirow{4}{*}{NUSD~\cite{wang2023non}} 
 & / & 0.514 (0.046) & 0.300 (0.137) & \textbf{0.745 (0.102)} \\
 & Mixup & 0.486 (0.052) &  0.357 (0.134) &  0.636 (0.156) \\
 & Instruct-TTS (CDoA)& 0.485 (0.038)&0.400 (0.193) & 0.606 (0.183) \\
 & DepFlow (CDoA) (Ours) &\textbf{0.577 (0.035)} & \textbf{0.471 (0.081)} & 0.697 (0.057)\\
\midrule
\multirow{5}{*}{\textit{HAREN}-CTC \cite{li2025hierarchical}} 
 & / & 0.525 (0.035) & 0.386 (0.081) & 0.673 (0.062) \\
 & SpecAugment & 0.477 (0.067) &  0.386 (0.041) &  0.600 (0.114) \\
 & Mixup & 0.526 (0.070) &  0.375 (0.090) &  \textbf{0.712 (0.072)} \\
 & Instruct-TTS (CDoA) & 0.510 (0.033)& \textbf{0.429 (0.033)} & 0.606 (0.033)\\
 & DepFlow (CDoA) (Ours) &   \textbf{0.551 (0.028)} & 0.400 (0.081) & 0.709 (0.090) \\
\bottomrule
\end{tabular}

\label{tab:result_main}
\end{table*}

\begin{table}[t]
\caption{Objective evaluation of TTS quality for natural and DepFlow-synthesized speech.
SIM-o is not reported for natural DAIC-WOZ speech because it represents the upper-bound self-similarity (100\%) by definition.}
\centering
\small
\begin{tabular}{cccc}
\toprule
Data Type & Severity & WER (\%)$\downarrow$ & SIM-o$\uparrow$ \\
\midrule
\multirow{6}{*}{DAIC-WOZ} & Healthy            & {13.76 (0.182)} & /  \\
& Mild             & {14.77 (0.176)} & /  \\
& Moderate             & {15.20 (0.175)} & /  \\
& Mod Severe             & {11.41 (0.193)} & /  \\
& Severe             & {15.15 (0.208)} & /  \\
& \textbf{Average} & \textbf{14.06 (0.150)} & / \\

\midrule
\multirow{6}{*}{Synth}& Healthy         & {13.94 (0.225)} & {57.13 (0.095)} \\
& Mild            & {14.03 (0.228)} & {57.07 (0.095)}\\
& Moderate              & {13.81 (0.227)} & {57.03 (0.095)}\\
& Mod Severe       & {13.93 (0.224)} & {56.88 (0.094)} \\
& Severe       & {13.92 (0.230)} & {56.73 (0.094)} \\
& \textbf{Average} & \textbf{13.93 (0.227)} & \textbf{56.97} (0.095)\\
\bottomrule
\end{tabular}
\label{tab:tts_obj}
\end{table}

\subsubsection{Intra-Speaker Acoustic Controllability}
We further examine controllability at the level of low-level acoustic markers. For each synthesized sample, a set of clinically relevant features is extracted, and intra-speaker Spearman correlations are computed between the intended severity levels and these features across the five synthesized variants.

Table~\ref{tab:dae_disentangle} summarizes the results. Several clinically recognized depression correlates exhibit strong and consistent monotonic trends:

\begin{itemize}
    \item The silence–speech ratio shows the strongest positive correlation, aligning with reports of reduced speech activity and psychomotor retardation in depression.
    \item Voice-quality measures such as shimmer and HNR change in the expected directions, suggesting increasingly unstable and breathier phonation as severity increases.
    \item Formant behavior (F1, F2, and centralization) shifts systematically with severity, reflecting vowel centralization and articulatory undershooting often observed in depressive speech.
\end{itemize}

These findings indicate that DepFlow does not merely produce perceptually different variants, but modulates speech along dimensions that are clinically meaningful and acoustically interpretable. Notably, $F_0$ statistics show weaker correlations, suggesting that DepFlow primarily controls spectral–articulatory and voice-quality cues rather than relying on pitch manipulation alone. Taken together with the embedding-level analyses, this demonstrates that DepFlow enables smooth, structured, and clinically grounded control over depression severity.

\subsubsection{TTS Quality and Speaker Similarity}
We next assess whether severity manipulation compromises linguistic intelligibility or speaker identity. Table~\ref{tab:tts_obj} reports Word Error Rate (WER) and speaker similarity (SIM-o) results for natural DAIC-WOZ utterances and DepFlow-synthesized speech across the five severity levels.

As shown in Table~\ref{tab:tts_obj}, the WER of synthesized speech closely matches that of natural recordings (13.93\% vs. 14.06\%), confirming that DepFlow preserves linguistic intelligibility despite the heavy manipulation of paralinguistic acoustic features.

Speaker similarity also remains stable across severity levels, suggesting that DepFlow primarily alters expressive and prosodic dimensions without distorting timbral identity. In other words, the imposed depressive cues modify how the subject speaks rather than who is speaking.

In sum, these results demonstrate that DepFlow can inject depression-related acoustic patterns in a controlled manner while maintaining speech quality and speaker characteristics, thereby avoiding artifacts that might otherwise confound downstream depression detection models.

\subsection{Effectiveness of CDoA Dataset for Downstream Depression Detection}

We next investigate whether DepFlow-generated CDoA samples improve downstream detection performance. These experiments aim to assess whether exposure to semantically incongruent samples leads models to develop more balanced and robust decision boundaries.

\subsubsection{Comparison with other Data Augmentation Methods}

Table~\ref{tab:result_main} presents the subject-level performance comparison against both conventional augmentation strategies (FrAUG~\cite{ravi2022fraug}, SpecAugment~\cite{park2019specaugment}, Mixup~\cite{zhang2017mixup}) and an Instruct-TTS baseline. DepFlow consistently outperforms all baselines across three distinct architectures. Notably, our method achieves superior performance compared to the CosyVoice-2–based \cite{du2024cosyvoice} voice-cloned TTS baseline, which also generates camouflage-oriented augmentation data but relies on generic voice cloning without learning a depression-aware acoustic manifold.

This comparison serves as a validation of our proposed framework: while generic TTS can produce diverse prosody, it fails to synthesize clinically meaningful depressive markers. The significant performance gap confirms that the robustness gains stem specifically from the targeted injection of depression-relevant acoustic cues, rather than mere data expansion.

\section{Conclusion}
In summary, DepFlow provides a waveform-level, depression-conditioned speech generation framework that constructs Camouflage Depression-oriented Augmentation dataset (CDoA dataset) by decoupling depressive acoustic cues from linguistic sentiment while preserving speech quality and speaker identity. These semantically incongruent yet clinically meaningful conditions are rarely represented in existing corpora, but are crucial for improving the robustness of downstream depression detection models through targeted data augmentation.

Empirical results demonstrate that DepFlow produces a coherent severity manifold, affords smooth and interpretable control of depressive acoustic cues, and preserves both speech intelligibility and speaker identity. When used as a augmentation source, DepFlow-generated CDoA dataset consistently improves the performance of multiple downstream depression detection architectures. While not establishing causal proof, the observed gains in sensitivity, specificity, macro-F1, and decision-boundary stability provide converging evidence that controlled acoustic–semantic mismatches reduce reliance on linguistic shortcuts and encourage models to attend to depression-related acoustic biomarkers. DepFlow therefore offers a principled way to leverage controllable generative modeling to regularize depression detectors and enhance robustness.

Several limitations remain. The learned severity axis, although internally coherent and aligned with known acoustic correlates of depression, has not yet been validated against clinician ratings or across broader populations and datasets. From an ethical perspective, the ability to synthesize speech with depressive acoustic characteristics introduces potential risks, including misuse for fabrication, unintended reinforcement of cultural stereotypes of depression, and psychological harm. Responsible use therefore requires transparent governance, informed consent, careful handling of sensitive speech data, and consideration of safeguards such as provenance tracking or watermarking. Future work will also explore perceptual studies of synthesized cues, extensions to multilingual and demographically diverse speakers, and broader deployment frameworks that balance robustness gains with reliability, interpretability, and societal impact in speech-based mental health assessment.

\bibliographystyle{unsrtnat}
\bibliography{references}  %%% Uncomment this line and comment out the ``thebibliography'' section below to use the external .bib file (using bibtex) .

@inproceedings{ma2016depaudionet,
  title={Depaudionet: An efficient deep model for audio based depression classification},
  author={Ma, Xingchen and Yang, Hongyu and Chen, Qiang and Huang, Di and Wang, Yunhong},
  booktitle={Proceedings of the 6th international workshop on audio/visual emotion challenge},
  pages={35--42},
  year={2016}
}

@article{chen2022speechformer,
  title={SpeechFormer: A hierarchical efficient framework incorporating the characteristics of speech},
  author={Chen, Weidong and Xing, Xiaofen and Xu, Xiangmin and Pang, Jianxin and Du, Lan},
  journal={arXiv preprint arXiv:2203.03812},
  year={2022}
}

@article{pearson1900x,
  title={X. On the criterion that a given system of deviations from the probable in the case of a correlated system of variables is such that it can be reasonably supposed to have arisen from random sampling},
  author={Pearson, Karl},
  journal={The London, Edinburgh, and Dublin Philosophical Magazine and Journal of Science},
  volume={50},
  number={302},
  pages={157--175},
  year={1900},
  publisher={Taylor \& Francis}
}

@article{zhang2017mixup,
  title={mixup: Beyond empirical risk minimization},
  author={Zhang, Hongyi and Cisse, Moustapha and Dauphin, Yann N and Lopez-Paz, David},
  journal={arXiv preprint arXiv:1710.09412},
  year={2017}
}

@inproceedings{kornblith2019similarity,
  title={Similarity of neural network representations revisited},
  author={Kornblith, Simon and Norouzi, Mohammad and Lee, Honglak and Hinton, Geoffrey},
  booktitle={International conference on machine learning},
  pages={3519--3529},
  year={2019},
  organization={PMlR}
}

@book{draper1998applied,
  title={Applied regression analysis},
  author={Draper, NR},
  year={1998},
  publisher={McGraw-Hill. Inc}
}

@article{hansen2015speaker,
  title={Speaker recognition by machines and humans: A tutorial review},
  author={Hansen, John HL and Hasan, Taufiq},
  journal={IEEE Signal processing magazine},
  volume={32},
  number={6},
  pages={74--99},
  year={2015},
  publisher={IEEE}
}

@article{cummins2015review,
  title={A review of depression and suicide risk assessment using speech analysis},
  author={Cummins, Nicholas and Scherer, Stefan and Krajewski, Jarek and Schnieder, Sebastian and Epps, Julien and Quatieri, Thomas F},
  journal={Speech communication},
  volume={71},
  pages={10--49},
  year={2015},
  publisher={Elsevier}
}

@article{mccorry2007physiology,
  title={Physiology of the autonomic nervous system},
  author={McCorry, Laurie Kelly},
  journal={American journal of pharmaceutical education},
  volume={71},
  number={4},
  pages={78},
  year={2007}
}

@article{low2010detection,
  title={Detection of clinical depression in adolescents’ speech during family interactions},
  author={Low, Lu-Shih Alex and Maddage, Namunu C and Lech, Margaret and Sheeber, Lisa B and Allen, Nicholas B},
  journal={IEEE transactions on biomedical engineering},
  volume={58},
  number={3},
  pages={574--586},
  year={2010},
  publisher={IEEE}
}

@article{almaghrabi2023bio,
  title={Bio-acoustic features of depression: A review},
  author={Almaghrabi, Shaykhah A and Clark, Scott R and Baumert, Mathias},
  journal={Biomedical Signal Processing and Control},
  volume={85},
  pages={105020},
  year={2023},
  publisher={Elsevier}
}

@article{lishi2025vector,
  title={Vector Quantization-based Counterfactual Augmentation for Speech-based Depression Detection under Data Scarcity},
  author={Lishi, ZUO and Mak, Man-Wai},
  journal={IEEE journal of biomedical and health informatics},
  year={2025},
  publisher={IEEE}
}

@article{morales2025mapping,
  title={Mapping the Neurophysiological Link Between Voice and Autonomic Function: A Scoping Review},
  author={Morales-Luque, Carmen and Carrillo-Franco, Laura and L{\'o}pez-Gonz{\'a}lez, Manuel V{\'\i}ctor and Gonz{\'a}lez-Garc{\'\i}a, Marta and Dawid-Milner, Marc Stefan},
  journal={Biology},
  volume={14},
  number={10},
  pages={1382},
  year={2025},
  publisher={MDPI}
}

@inproceedings{shoemake1985animating,
  title={Animating rotation with quaternion curves},
  author={Shoemake, Ken},
  booktitle={Proceedings of the 12th annual conference on Computer graphics and interactive techniques},
  pages={245--254},
  year={1985}
}

@article{scherer2003vocal,
  title={Vocal communication of emotion: A review of research paradigms},
  author={Scherer, Klaus R},
  journal={Speech communication},
  volume={40},
  number={1-2},
  pages={227--256},
  year={2003},
  publisher={Elsevier}
}

@inproceedings{ringeval2019avec,
  title={AVEC 2019 workshop and challenge: state-of-mind, detecting depression with AI, and cross-cultural affect recognition},
  author={Ringeval, Fabien and Schuller, Bj{\"o}rn and Valstar, Michel and Cummins, Nicholas and Cowie, Roddy and Tavabi, Leili and Schmitt, Maximilian and Alisamir, Sina and Amiriparian, Shahin and Messner, Eva-Maria and others},
  booktitle={Proceedings of the 9th International on Audio/visual Emotion Challenge and Workshop},
  pages={3--12},
  year={2019}
}

@inproceedings{valstar2013avec,
  title={Avec 2013: the continuous audio/visual emotion and depression recognition challenge},
  author={Valstar, Michel and Schuller, Bj{\"o}rn and Smith, Kirsty and Eyben, Florian and Jiang, Bihan and Bilakhia, Sanjay and Schnieder, Sebastian and Cowie, Roddy and Pantic, Maja},
  booktitle={Proceedings of the 3rd ACM international workshop on Audio/visual emotion challenge},
  pages={3--10},
  year={2013}
}

@article{bandyopadhyay2014american,
  title={American Psychiatric Association, Diagnostic and Statistical Manual of Mental Disorders: Dsm-5, Washington, DC, American Psychiatric Publishing, 2013. Ananth Mahesh, In defense of an evolutionary concept of health nature, norms, and human biology, Aldershot, England, Ashgate},
  author={Bandyopadhyay Prasanta, S and Forster Malcolm, R and Oxford, Elsevier and Barkow Jerome, H and Leda, Cosmides and John, Tooby and William, Bechtel and Richardson Robert, C and Beck Aaron, T and John, Rush A and others},
  journal={Philosophy},
  volume={39},
  number={6},
  pages={683--724},
  year={2014}
}

@article{lipman2022flow,
  title={Flow matching for generative modeling},
  author={Lipman, Yaron and Chen, Ricky TQ and Ben-Hamu, Heli and Nickel, Maximilian and Le, Matt},
  journal={arXiv preprint arXiv:2210.02747},
  year={2022}
}

@book{pearl2009causality,
  title={Causality},
  author={Pearl, Judea},
  year={2009},
  publisher={Cambridge university press}
}

@article{scholkopf2021toward,
  title={Toward causal representation learning},
  author={Sch{\"o}lkopf, Bernhard and Locatello, Francesco and Bauer, Stefan and Ke, Nan Rosemary and Kalchbrenner, Nal and Goyal, Anirudh and Bengio, Yoshua},
  journal={Proceedings of the IEEE},
  volume={109},
  number={5},
  pages={612--634},
  year={2021},
  publisher={IEEE}
}

@article{loshchilov2017decoupled,
  title={Decoupled weight decay regularization},
  author={Loshchilov, Ilya and Hutter, Frank},
  journal={arXiv preprint arXiv:1711.05101},
  year={2017}
}

@inproceedings{wang2023non,
  title={Non-uniform speaker disentanglement for depression detection from raw speech signals},
  author={Wang, Jinhan and Ravi, Vijay and Alwan, Abeer},
  booktitle={Interspeech},
  volume={2023},
  pages={2343},
  year={2023}
}

@book{jansson2016psychiatric,
  title={The psychiatric interview for differential diagnosis},
  author={Jansson, Lennart and Nordgaard, Julie and others},
  volume={270},
  year={2016},
  publisher={Springer}
}

@article{shetty2018understanding,
  title={Understanding masked depression: A Clinical scenario},
  author={Shetty, Prasad and Mane, Akshata and Fulmali, Sourabh and Uchit, Ganesh},
  journal={Indian journal of psychiatry},
  volume={60},
  number={1},
  pages={97--102},
  year={2018},
  publisher={Medknow}
}

@inproceedings{gratch2014distress,
  title={The distress analysis interview corpus of human and computer interviews.},
  author={Gratch, Jonathan and Artstein, Ron and Lucas, Gale M and Stratou, Giota and Scherer, Stefan and Nazarian, Angela and Wood, Rachel and Boberg, Jill and DeVault, David and Marsella, Stacy and others},
  booktitle={LREC},
  pages={3123--3128},
  year={2014},
  organization={Reykjavik}
}

@inproceedings{ravi2022step,
  title={A step towards preserving speakers’ identity while detecting depression via speaker disentanglement},
  author={Ravi, Vijay and Wang, Jinhan and Flint, Jonathan and Alwan, Abeer},
  booktitle={Interspeech},
  volume={2022},
  pages={3338},
  year={2022}
}

@inproceedings{ravi2022fraug,
  title={Fraug: A frame rate based data augmentation method for depression detection from speech signals},
  author={Ravi, Vijay and Wang, Jinhan and Flint, Jonathan and Alwan, Abeer},
  booktitle={ICASSP 2022-2022 IEEE International Conference on Acoustics, Speech and Signal Processing (ICASSP)},
  pages={6267--6271},
  year={2022},
  organization={IEEE}
}

@article{goodfellow2020generative,
  title={Generative adversarial networks},
  author={Goodfellow, Ian and Pouget-Abadie, Jean and Mirza, Mehdi and Xu, Bing and Warde-Farley, David and Ozair, Sherjil and Courville, Aaron and Bengio, Yoshua},
  journal={Communications of the ACM},
  volume={63},
  number={11},
  pages={139--144},
  year={2020},
  publisher={ACM New York, NY, USA}
}

@article{yang2020feature,
  title={Feature augmenting networks for improving depression severity estimation from speech signals},
  author={Yang, Le and Jiang, Dongmei and Sahli, Hichem},
  journal={IEEE Access},
  volume={8},
  pages={24033--24045},
  year={2020},
  publisher={IEEE}
}

@inproceedings{feng2023knowledge,
  title={A knowledge-driven vowel-based approach of depression classification from speech using data augmentation},
  author={Feng, Kexin and Chaspari, Theodora},
  booktitle={ICASSP 2023-2023 IEEE International Conference on Acoustics, Speech and Signal Processing (ICASSP)},
  pages={1--5},
  year={2023},
  organization={IEEE}
}

@inproceedings{kumar2024depression,
  title={Depression Classification Using Log-Mel Spectrograms: A Comparative Analysis of Window Size-Based Data Augmentation and Deep Learning Models},
  author={Kumar, Lokesh and Kaustubh, Kumar and Mattur, Shashaank Aswatha and Prasanna, SR Mahadeva},
  booktitle={2024 27th Conference of the Oriental COCOSDA International Committee for the Co-ordination and Standardisation of Speech Databases and Assessment Techniques (O-COCOSDA)},
  pages={1--6},
  year={2024},
  organization={IEEE}
}

@article{li2025automated,
  title={Automated Depression Detection from Text and Audio: A Systematic Review},
  author={Li, Yuxin and Kumbale, Sinchana and Chen, Yanru and Surana, Tanmay and Chng, Eng Siong and Guan, Cuntai},
  journal={IEEE Journal of Biomedical and Health Informatics},
  year={2025},
  publisher={IEEE}
}

@article{chlasta2019automated,
  title={Automated speech-based screening of depression using deep convolutional neural networks},
  author={Chlasta, Karol and Wo{\l}k, Krzysztof and Krejtz, Izabela},
  journal={Procedia Computer Science},
  volume={164},
  pages={618--628},
  year={2019},
  publisher={Elsevier}
}

@inproceedings{wu2023self,
  title={Self-supervised representations in speech-based depression detection},
  author={Wu, Wen and Zhang, Chao and Woodland, Philip C},
  booktitle={ICASSP 2023-2023 IEEE International Conference on Acoustics, Speech and Signal Processing (ICASSP)},
  pages={1--5},
  year={2023},
  organization={IEEE}
}

@inproceedings{liang2025depressgen,
  title={DepressGEN: Synthetic Data Generation Framework for Depression Detection},
  author={Liang, Wenrui and Zhang, Rong and Zhang, Xuezhen and Ma, Ying and Zhang, Wei-Qiang},
  booktitle={Proc. Interspeech 2025},
  pages={464--468},
  year={2025}
}

@inproceedings{saidi2020hybrid,
  title={Hybrid CNN-SVM classifier for efficient depression detection system},
  author={Saidi, Afef and Othman, Slim Ben and Saoud, Slim Ben},
  booktitle={2020 4th International Conference on Advanced Systems and Emergent Technologies (IC\_ASET)},
  pages={229--234},
  year={2020},
  organization={IEEE}
}

@article{geirhos2020shortcut,
  title={Shortcut learning in deep neural networks},
  author={Geirhos, Robert and Jacobsen, J{\"o}rn-Henrik and Michaelis, Claudio and Zemel, Richard and Brendel, Wieland and Bethge, Matthias and Wichmann, Felix A},
  journal={Nature Machine Intelligence},
  volume={2},
  number={11},
  pages={665--673},
  year={2020},
  publisher={Nature Publishing Group UK London}
}

@inproceedings{dubagunta2019learning,
  title={Learning voice source related information for depression detection},
  author={Dubagunta, S Pavankumar and Vlasenko, Bogdan and Doss, Mathew Magimai-},
  booktitle={ICASSP 2019-2019 IEEE International Conference on Acoustics, Speech and Signal Processing (ICASSP)},
  pages={6525--6529},
  year={2019},
  organization={IEEE}
}

@inproceedings{Othmani2021,
  title={Towards robust deep neural networks for affect and depression recognition from speech},
  author={Othmani, Alice and Kadoch, Daoud and Bentounes, Kamil and Rejaibi, Emna and Alfred, Romain and Hadid, Abdenour},
  booktitle={Pattern Recognition. ICPR International Workshops and Challenges: Virtual Event, January 10--15, 2021, Proceedings, Part II},
  pages={5--19},
  year={2021},
  organization={Springer}
}

@article{vazquez2020automatic,
  title={Automatic detection of depression in speech using ensemble convolutional neural networks},
  author={V{\'a}zquez-Romero, Adri{\'a}n and Gallardo-Antol{\'\i}n, Ascensi{\'o}n},
  journal={Entropy},
  volume={22},
  number={6},
  pages={688},
  year={2020},
  publisher={MDPI}
}

@inproceedings{wang2022ecapa,
  title={ECAPA-TDNN Based Depression Detection from Clinical Speech.},
  author={Wang, Dong and Ding, Yanhui and Zhao, Qing and Yang, Peilin and Tan, Shuping and Li, Ya},
  booktitle={Interspeech},
  pages={3333--3337},
  year={2022}
}

@article{Salekin2018,
  title={A weakly supervised learning framework for detecting social anxiety and depression},
  author={Salekin, Asif and Eberle, Jeremy W and Glenn, Jeffrey J and Teachman, Bethany A and Stankovic, John A},
  journal={Proceedings of the ACM on interactive, mobile, wearable and ubiquitous technologies},
  volume={2},
  number={2},
  pages={1--26},
  year={2018},
  publisher={ACM New York, NY, USA}
}

@inproceedings{zhao2020hierarchical,
  title={Hierarchical attention transfer networks for depression assessment from speech},
  author={Zhao, Ziping and Bao, Zhongtian and Zhang, Zixing and Cummins, Nicholas and Wang, Haishuai and Schuller, Bj{\"o}rn},
  booktitle={ICASSP 2020-2020 IEEE international conference on acoustics, speech and signal processing (ICASSP)},
  pages={7159--7163},
  year={2020},
  organization={IEEE}
}

@article{muzammel2020audvowelconsnet,
  title={AudVowelConsNet: A phoneme-level based deep CNN architecture for clinical depression diagnosis},
  author={Muzammel, Muhammad and Salam, Hanan and Hoffmann, Yann and Chetouani, Mohamed and Othmani, Alice},
  journal={Machine Learning with Applications},
  volume={2},
  pages={100005},
  year={2020},
  publisher={Elsevier}
}

@article{baevski2020wav2vec,
  title={wav2vec 2.0: A framework for self-supervised learning of speech representations},
  author={Baevski, Alexei and Zhou, Yuhao and Mohamed, Abdelrahman and Auli, Michael},
  journal={Advances in neural information processing systems},
  volume={33},
  pages={12449--12460},
  year={2020}
}

@article{hsu2021hubert,
  title={Hubert: Self-supervised speech representation learning by masked prediction of hidden units},
  author={Hsu, Wei-Ning and Bolte, Benjamin and Tsai, Yao-Hung Hubert and Lakhotia, Kushal and Salakhutdinov, Ruslan and Mohamed, Abdelrahman},
  journal={IEEE/ACM transactions on audio, speech, and language processing},
  volume={29},
  pages={3451--3460},
  year={2021},
  publisher={IEEE}
}

@article{chen2022wavlm,
  title={Wavlm: Large-scale self-supervised pre-training for full stack speech processing},
  author={Chen, Sanyuan and Wang, Chengyi and Chen, Zhengyang and Wu, Yu and Liu, Shujie and Chen, Zhuo and Li, Jinyu and Kanda, Naoyuki and Yoshioka, Takuya and Xiao, Xiong and others},
  journal={IEEE Journal of Selected Topics in Signal Processing},
  volume={16},
  number={6},
  pages={1505--1518},
  year={2022},
  publisher={IEEE}
}

@inproceedings{radford2023robust,
  title={Robust speech recognition via large-scale weak supervision},
  author={Radford, Alec and Kim, Jong Wook and Xu, Tao and Brockman, Greg and McLeavey, Christine and Sutskever, Ilya},
  booktitle={International conference on machine learning},
  pages={28492--28518},
  year={2023},
  organization={PMLR}
}

@inproceedings{Zhang2021Depa,
  title={Depa: Self-supervised audio embedding for depression detection},
  author={Zhang, Pingyue and Wu, Mengyue and Dinkel, Heinrich and Yu, Kai},
  booktitle={Proceedings of the 29th ACM international conference on multimedia},
  pages={135--143},
  year={2021}
}

@inproceedings{Toto2021,
  title={Audibert: A deep transfer learning multimodal classification framework for depression screening},
  author={Toto, Ermal and Tlachac, ML and Rundensteiner, Elke A},
  booktitle={Proceedings of the 30th ACM international conference on information \& knowledge management},
  pages={4145--4154},
  year={2021}
}

@article{li2025hierarchical,
  title={Hierarchical Self-Supervised Representation Learning for Depression Detection from Speech},
  author={Li, Yuxin and Chng, Eng Siong and Guan, Cuntai},
  journal={arXiv preprint arXiv:2510.08593},
  year={2025}
}

@inproceedings{mehta2024matcha,
  title={Matcha-TTS: A fast TTS architecture with conditional flow matching},
  author={Mehta, Shivam and Tu, Ruibo and Beskow, Jonas and Sz{\'e}kely, {\'E}va and Henter, Gustav Eje},
  booktitle={ICASSP 2024-2024 IEEE International Conference on Acoustics, Speech and Signal Processing (ICASSP)},
  pages={11341--11345},
  year={2024},
  organization={IEEE}
}

@inproceedings{perez2018film,
  title={Film: Visual reasoning with a general conditioning layer},
  author={Perez, Ethan and Strub, Florian and De Vries, Harm and Dumoulin, Vincent and Courville, Aaron},
  booktitle={Proceedings of the AAAI conference on artificial intelligence},
  volume={32},
  number={1},
  year={2018}
}

@article{le2023voicebox,
  title={Voicebox: Text-guided multilingual universal speech generation at scale},
  author={Le, Matthew and Vyas, Apoorv and Shi, Bowen and Karrer, Brian and Sari, Leda and Moritz, Rashel and Williamson, Mary and Manohar, Vimal and Adi, Yossi and Mahadeokar, Jay and others},
  journal={Advances in neural information processing systems},
  volume={36},
  pages={14005--14034},
  year={2023}
}

@article{kroenke2009phq,
  title={The PHQ-8 as a measure of current depression in the general population},
  author={Kroenke, Kurt and Strine, Tara W and Spitzer, Robert L and Williams, Janet BW and Berry, Joyce T and Mokdad, Ali H},
  journal={Journal of affective disorders},
  volume={114},
  number={1-3},
  pages={163--173},
  year={2009},
  publisher={Elsevier}
}

@article{anastassiou2024seed,
  title={Seed-tts: A family of high-quality versatile speech generation models},
  author={Anastassiou, Philip and Chen, Jiawei and Chen, Jitong and Chen, Yuanzhe and Chen, Zhuo and Chen, Ziyi and Cong, Jian and Deng, Lelai and Ding, Chuang and Gao, Lu and others},
  journal={arXiv preprint arXiv:2406.02430},
  year={2024}
}

@inproceedings{chen2022large,
  title={Large-scale self-supervised speech representation learning for automatic speaker verification},
  author={Chen, Zhengyang and Chen, Sanyuan and Wu, Yu and Qian, Yao and Wang, Chengyi and Liu, Shujie and Qian, Yanmin and Zeng, Michael},
  booktitle={ICASSP 2022-2022 IEEE International Conference on Acoustics, Speech and Signal Processing (ICASSP)},
  pages={6147--6151},
  year={2022},
  organization={IEEE}
}

@article{desplanques2020ecapa,
  title={Ecapa-tdnn: Emphasized channel attention, propagation and aggregation in tdnn based speaker verification},
  author={Desplanques, Brecht and Thienpondt, Jenthe and Demuynck, Kris},
  journal={arXiv preprint arXiv:2005.07143},
  year={2020}
}

@article{ganin2016domain,
  title={Domain-adversarial training of neural networks},
  author={Ganin, Yaroslav and Ustinova, Evgeniya and Ajakan, Hana and Germain, Pascal and Larochelle, Hugo and Laviolette, Fran{\c{c}}ois and March, Mario and Lempitsky, Victor},
  journal={Journal of machine learning research},
  volume={17},
  number={59},
  pages={1--35},
  year={2016}
}

@article{brown2025camouflaging,
  title={Camouflaging depression},
  author={Brown, Seth},
  journal={Discover Mental Health},
  volume={5},
  number={1},
  pages={71},
  year={2025},
  publisher={Springer}
}

@article{yamagishi2019cstr,
  title={CSTR VCTK Corpus: English multi-speaker corpus for CSTR voice cloning toolkit (version 0.92)},
  author={Yamagishi, Junichi and Veaux, Christophe and MacDonald, Kirsten},
  journal={The Rainbow Passage which the speakers read out can be found in the International Dialects of English Archive:(http://web. ku. edu/\~{} idea/readings/rainbow. htm).},
  year={2019}
}

@article{zhang2023speechgpt,
  title={Speechgpt: Empowering large language models with intrinsic cross-modal conversational abilities},
  author={Zhang, Dong and Li, Shimin and Zhang, Xin and Zhan, Jun and Wang, Pengyu and Zhou, Yaqian and Qiu, Xipeng},
  journal={arXiv preprint arXiv:2305.11000},
  year={2023}
}

@article{du2024cosyvoice,
  title={Cosyvoice 2: Scalable streaming speech synthesis with large language models},
  author={Du, Zhihao and Wang, Yuxuan and Chen, Qian and Shi, Xian and Lv, Xiang and Zhao, Tianyu and Gao, Zhifu and Yang, Yexin and Gao, Changfeng and Wang, Hui and others},
  journal={arXiv preprint arXiv:2412.10117},
  year={2024}
}

@article{park2019specaugment,
  title={Specaugment: A simple data augmentation method for automatic speech recognition},
  author={Park, Daniel S and Chan, William and Zhang, Yu and Chiu, Chung-Cheng and Zoph, Barret and Cubuk, Ekin D and Le, Quoc V},
  journal={arXiv preprint arXiv:1904.08779},
  year={2019}
}

@article{ji2024wavchat,
  title={Wavchat: A survey of spoken dialogue models},
  author={Ji, Shengpeng and Chen, Yifu and Fang, Minghui and Zuo, Jialong and Lu, Jingyu and Wang, Hanting and Jiang, Ziyue and Zhou, Long and Liu, Shujie and Cheng, Xize and others},
  journal={arXiv preprint arXiv:2411.13577},
  year={2024}
}

@article{guo2025deepseek,
  title={Deepseek-r1: Incentivizing reasoning capability in llms via reinforcement learning},
  author={Guo, Daya and Yang, Dejian and Zhang, Haowei and Song, Junxiao and Zhang, Ruoyu and Xu, Runxin and Zhu, Qihao and Ma, Shirong and Wang, Peiyi and Bi, Xiao and others},
  journal={arXiv preprint arXiv:2501.12948},
  year={2025}
}

@article{harrell1982evaluating,
  title={Evaluating the yield of medical tests},
  author={Harrell, Frank E and Califf, Robert M and Pryor, David B and Lee, Kerry L and Rosati, Robert A},
  journal={Jama},
  volume={247},
  number={18},
  pages={2543--2546},
  year={1982},
  publisher={American Medical Association}
}

@article{spearman1987proof,
  title={The proof and measurement of association between two things},
  author={Spearman, Charles},
  journal={The American journal of psychology},
  volume={100},
  number={3/4},
  pages={441--471},
  year={1987},
  publisher={JSTOR}
}

@article{sokolova2009systematic,
  title={A systematic analysis of performance measures for classification tasks},
  author={Sokolova, Marina and Lapalme, Guy},
  journal={Information processing \& management},
  volume={45},
  number={4},
  pages={427--437},
  year={2009},
  publisher={Elsevier}
}

@inproceedings{bailey2021gender,
  title={Gender bias in depression detection using audio features},
  author={Bailey, Andrew and Plumbley, Mark D},
  booktitle={2021 29th European Signal Processing Conference (EUSIPCO)},
  pages={596--600},
  year={2021},
  organization={IEEE}
}

@article{kong2020hifi,
  title={Hifi-gan: Generative adversarial networks for efficient and high fidelity speech synthesis},
  author={Kong, Jungil and Kim, Jaehyeon and Bae, Jaekyoung},
  journal={Advances in neural information processing systems},
  volume={33},
  pages={17022--17033},
  year={2020}
}

@article{zhang2025speecht,
  title={SpeechT-RAG: Reliable Depression Detection in LLMs with Retrieval-Augmented Generation Using Speech Timing Information},
  author={Zhang, Xiangyu and Liu, Hexin and Zhang, Qiquan and Ahmed, Beena and Epps, Julien},
  journal={arXiv preprint arXiv:2502.10950},
  year={2025}
}

@inproceedings{zhang2024speaking,
  title={Speaking in wavelet domain: A simple and efficient approach to speed up speech diffusion model},
  author={Zhang, Xiangyu and Liu, Daijiao and Liu, Hexin and Zhang, Qiquan and Meng, Hanyu and Perera, Leibny Paola Garcia and Chng, EngSiong and Yao, Lina},
  booktitle={Proceedings of the 2024 Conference on Empirical Methods in Natural Language Processing},
  pages={159--171},
  year={2024}
}




% 	\bibitem{keshet2016prediction}
% 	Keshet, Renato, Alina Maor, and George Kour.
% 	\newblock Prediction-Based, Prioritized Market-Share Insight Extraction.
% 	\newblock In {\em Advanced Data Mining and Applications (ADMA), 2016 12th International 
%                       Conference of}, pages 81--94,2016.

% \end{thebibliography}

\end{document}